\pdfoutput=1

\documentclass[11pt]{article}

\usepackage[final]{acl}

\usepackage{times}
\usepackage{latexsym}

\usepackage[T1]{fontenc}

\usepackage[utf8]{inputenc}

\usepackage{microtype}

\usepackage{inconsolata}

\usepackage{graphicx}
\usepackage{diagbox}
\usepackage{multirow}
\usepackage{booktabs}
\usepackage{amsmath}
\usepackage{lipsum}

\title{
    G2S: A General-to-Specific Learning Framework for Temporal Knowledge Graph Forecasting with Large Language Models
}

\author{
  Long Bai\textsuperscript{1,2,4 *}, Zixuan Li\textsuperscript{1,2}, Xiaolong Jin\textsuperscript{1,2,3 *},
\\
  \textbf{Jiafeng Guo\textsuperscript{1,2,3}, Xueqi Cheng\textsuperscript{1,2,3}, Tat-Seng Chua\textsuperscript{4}}
\\
  \textsuperscript{1}Key Laboratory of Network Data Science and Technology,
\\
  Institute of Computing Technology, Chinese Academy of Sciences
\\
  \textsuperscript{2}State Key Laboratory of AI Safety
\\
  \textsuperscript{3}School of Computer Science, University of Chinese Academy of Sciences
\\
  \textsuperscript{4}National University of Singapore
\\
  \texttt{\{bailong,jinxiaolong\}@ict.ac.cn}
}

\begin{document}
\maketitle

\newcommand\blfootnote[1]{%
  \begingroup
  \renewcommand\thefootnote{}\footnote{#1}
  \addtocounter{footnote}{-1}%
  \endgroup
}

\blfootnote{* Corresponding author}

\begin{abstract}
Forecasting over Temporal Knowledge Graphs (TKGs) which predicts future facts based on historical ones has received much attention.
Recent studies have introduced Large Language Models (LLMs) for this task to enhance the models' generalization abilities.
However, these models perform forecasting via simultaneously learning two kinds of entangled knowledge in the TKG:
(1) general patterns, i.e., invariant temporal structures shared across different scenarios;
and (2) scenario information, i.e., factual knowledge engaged in specific scenario, such as entities and relations.
As a result, the learning processes of these two kinds of knowledge may interfere with each other,
which potentially impact the generalization abilities of the models.
To enhance the generalization ability of LLMs on this task, in this paper, 
we propose a General-to-Specific learning framework (G2S) that disentangles the learning processes of the above two kinds of knowledge.
In the general learning stage, we mask the scenario information in different TKGs and convert it into anonymous temporal structures.
After training on these structures, the model is able to capture the general patterns across different TKGs.
In the specific learning stage, we inject the scenario information into the structures via either in-context learning or fine-tuning modes.
Experimental results show that G2S effectively improves the generalization abilities of LLMs.

\end{abstract}

\section{Introduction}
\label{sec:introduction}

Temporal Knowledge Graph (TKG), which consists of facts in the form of \textit{(subject, relation, object, timestamp)},
is often used to represent the changes of facts over time.
Forecasting future facts based on historical ones in TKG is critical for many time-sensitive scenarios,
such as finance~\cite{feng2019temporal}, healthcare~\cite{ernst2014knowlife}, and politics~\cite{ijcai2019p955}.
These scenarios involve different entities, relations, and time granularities,
which drive us to seek a model that is able to adapt to new scenarios efficiently and effectively.

Previous methods view TKG as a temporally ordered sequence of knowledge graphs and adopt graph neural networks~\cite{li-etal-2021-temporal} or recurrent neural networks~\cite{jin-etal-2020-recurrent} to model it.
These methods mainly focus on fitting a model to each specific TKG and overlook the generalization abilities.
Recent methods attempt to model different TKGs via a unified Large Language Model (LLM),
which shows considerable improvement in generalization ability.
Some methods append historical facts into the prompt and utilize the in-context learning technique to predict the answer in unseen scenarios~\cite{lee-etal-2023-temporal,xia2024chainofhistoryreasoningtemporalknowledge}.
Other methods learn temporal knowledge via supervised fine-tuning LLMs on a certain dataset and try to transfer this knowledge to unseen scenarios~\cite{liao-etal-2024-gentkg,ding-etal-2024-zrllm}.

\begin{figure}[t]
  \includegraphics[width=\columnwidth]{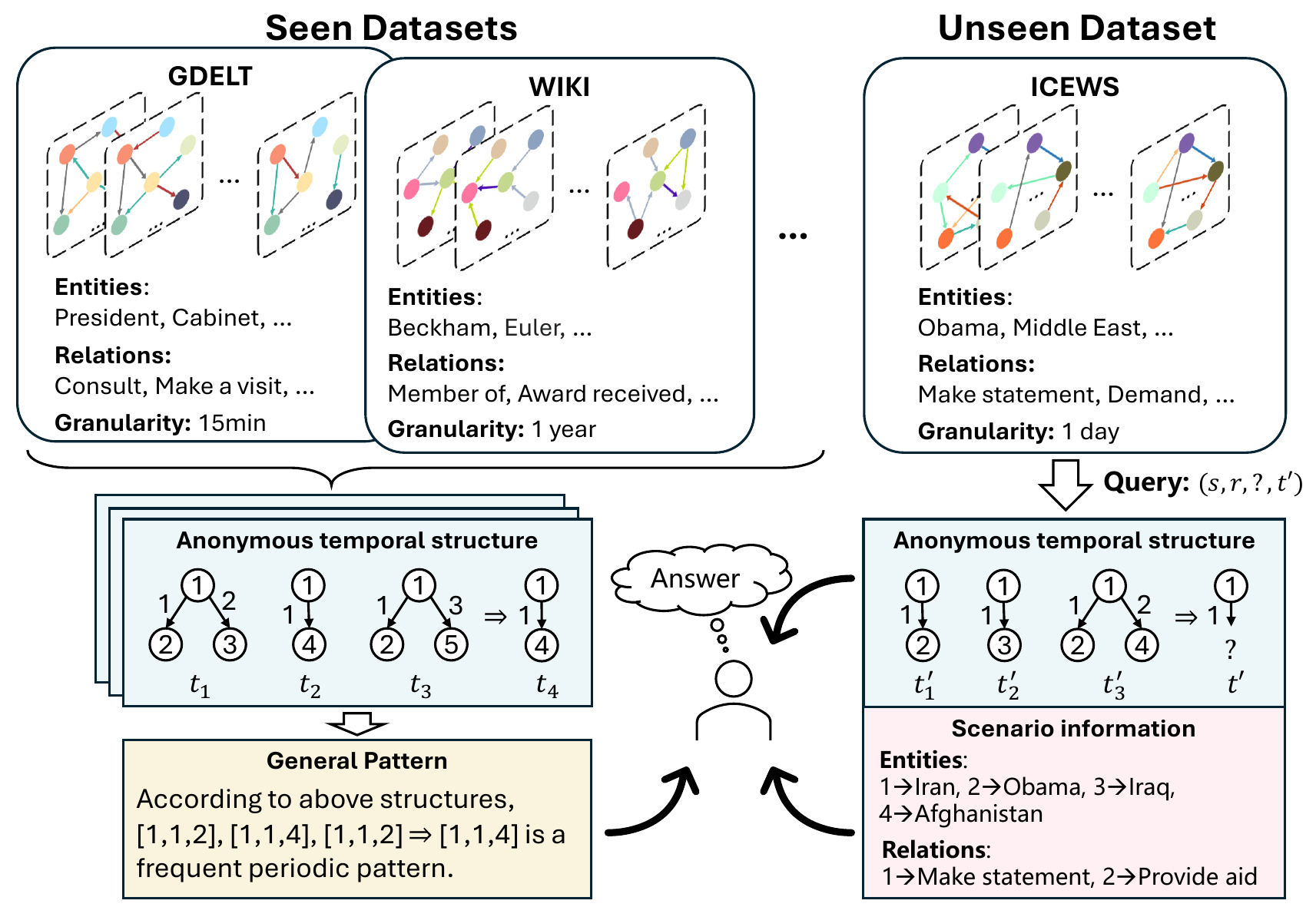}
  \caption{An example of learning general patterns and applying to unseen datasets.}
  \label{fig:intro}
  \vspace{-5mm}
\end{figure}

The generalization abilities of the TKG forecasting models mainly come from their abilities to learn general knowledge from seen scenarios and apply it to unseen scenarios.
However, different TKGs describe different scenarios that involve different entities, relations, and time granularities.
Taking Figure~\ref{fig:intro} as an example, GDELT is about conflicts and crisis scenarios in the geopolitics domain,
while WIKI contains sports and academic scenarios in the general domain.
This leads us to imagine: what is the general knowledge that can be shared across the scenarios in these TKGs?
We observe that, if we anonymize the specific entity and relation names into abstract IDs,
some temporal structures in different scenarios reveal a common substructure.
Following this observation, we think that the TKGs contain two kinds of entangled knowledge,
namely, general patterns and scenario information.
The general patterns are common structures that can be generalized to different scenarios,
while scenario information is factual knowledge, such as specific entities and relations,
which only help TKG forecasting in a certain scenario.
If the models directly learn the TKG (i.e., learn these two kinds of knowledge simultaneously),
their learning processes may interfere with each other.
From this point of view, previous methods, even LLM-based ones, may not learn the general patterns comprehensively, 
which limits their generalization abilities.

Motivated by this, in this paper, we aim to disentangle the above two kinds of knowledge in TKG,
and propose a general-to-specific learning framework for TKG forecasting, named G2S.
G2S consists of two stages, namely, the general learning stage and the specific learning stage.
In the general learning stage, 
we convert multiple TKGs into anonymous temporal structures via anonymizing the specific entities, relations, and timestamps into abstract IDs.
After being trained on these structures, the model is able to learn the general patterns across multiple TKGs comprehensively.
In the specific learning stage,
we split the unseen TKG into anonymous temporal structures and their mappings between the specific entities/relations and the abstract IDs,
as shown in the bottom right of Figure~\ref{fig:intro}.
In addition to general patterns, the model further learns the above information in either the in-context learning mode or the fine-tuning mode.
To verify the effectiveness of G2S, we evaluate the proposed framework under three settings,
namely, zero-shot, low-resource, and standard learning.
Experimental results show that G2S effectively improves the generalization abilities of LLMs.

The contributions of this paper are three-fold:
\begin{itemize}
    \item
        We identify the two kinds of entangled knowledge in TKG, 
        namely, the general patterns and scenario information,
        and disentangle them via several anonymization strategies.
    \item
        We propose a general-to-specific learning framework for TKG forecasting, 
        which prevents the learning processes of general patterns and scenario information from interfering with each other.
        In this way, the generalization ability of the model is enhanced.
    \item
        We conduct extensive experiments under three different settings.
        The experimental results demonstrate the effectiveness of the proposed framework.
\end{itemize}

\section{Related Work}
There are two different task settings for TKG reasoning,
namely, interpolation and extrapolation settings.
TKG reasoning under interpolation setting, also known as TKG completion,
aims to predict the missing entities at past timestamps~\cite{jiang-etal-2016-encoding},
while TKG reasoning under extrapolation setting, also known as TKG forecasting,
aims to predict the missing entities at future timestamps~\cite{jin-etal-2020-recurrent}.
This paper focuses on the latter one.

Traditional TKG forecasting methods can be categorized into three different types:
(1) \textbf{Neural-based methods} utilize deep neural networks to model the historical facts.
RE-Net~\cite{jin-etal-2020-recurrent} develops a recurrent neural network to model query related historical facts.
RE-GCN~\cite{li-etal-2021-temporal} utilizes graph convolutional networks to model recent KG snapshots.
TANGO~\cite{han-etal-2021-learning-neural} adopts neural ordinary differential equations to model the structure information among entities.
xERTE~\cite{han2021explainable} designs a subgraph expansion and pruning mechanism to find explainable supports for forecasting. 
(2) \textbf{Reinforcement learning-based methods} utilize reinforcement learning to find a path between query entity and answer entity,
such as CluSTeR~\cite{li-etal-2021-search} and TITer~\cite{sun-etal-2021-timetraveler}.
(3) \textbf{Rule-based methods} mine logical rules to derive the answer from historical facts, such as TLogic~\cite{Liu_Ma_Hildebrandt_Joblin_Tresp_2022}.
These methods mainly focus on enhancing the performance of the models on each specific TKG and overlook the generalization ability of the models.

Recently, some studies have introduced LLM for TKG forecasting,
which can be categorized into in-context learning and supervised fine-tuning methods.
\citet{lee-etal-2023-temporal} propose an in-context learning method,
which provides a basic input format to fit TKG data to LLMs.
\citet{xia2024chainofhistoryreasoningtemporalknowledge} investigate the LLM's ability to find useful historical facts and reasoning answers based on them.
GenTKG~\cite{liao-etal-2024-gentkg} proposes to select historical facts via rules and trains LLMs to answer the query via supervised fine-tuning.
These methods show generalization abilities on TKG forecasting to some extent.
However, as analyzed in Section~\ref{sec:introduction}, 
their learning processes of general patterns and scenario information may interfere with each other,
which limits their generalization abilities.

\section{Problem Formulation}
A TKG $\mathcal{G}$ is formalized as a sequence of KG snapshots from timestamps $1$ to $T$, i.e.,
$\mathcal{G}=\{\mathcal{G}_1, \mathcal{G}_2, ..., \mathcal{G}_T\}$.
Each KG snapshot $\mathcal{G}_t=\{\mathcal{E}, \mathcal{R}, \mathcal{F}_t\}$ consists of the entity set $\mathcal{E}$, relation set $\mathcal{R}$, and the fact set $\mathcal{F}_t$ at timestamp $t \in \{1,...,T\}$.
Each fact $f_i \in \mathcal{F}_t$ is denoted as a quadruple $f_i = (s_i, r_i, o_i, t)$,
where $s_i, o_i \in \mathcal{E}$ are subject and object entities; $r_i \in \mathcal{R}$ is the relation.
Currently, the TKG forecasting task focuses on predicting the missing object entity in a query $q=(s_q, r_q, ?, t_q)$.
The model $\mathcal{M}$ is required to score possible object entities based on the query and historical TKG snapshots: $\text{score}(o) = \mathcal{M}(q, \mathcal{G}_{<t_q})$. 
Note that, when predicting the missing subject entity, 
the query is formulated as $q=(?, r_q, o_q, t_q)$.

\section{Methodology}
\begin{figure*}[t]
  \centering
  \includegraphics[width=0.9\linewidth]{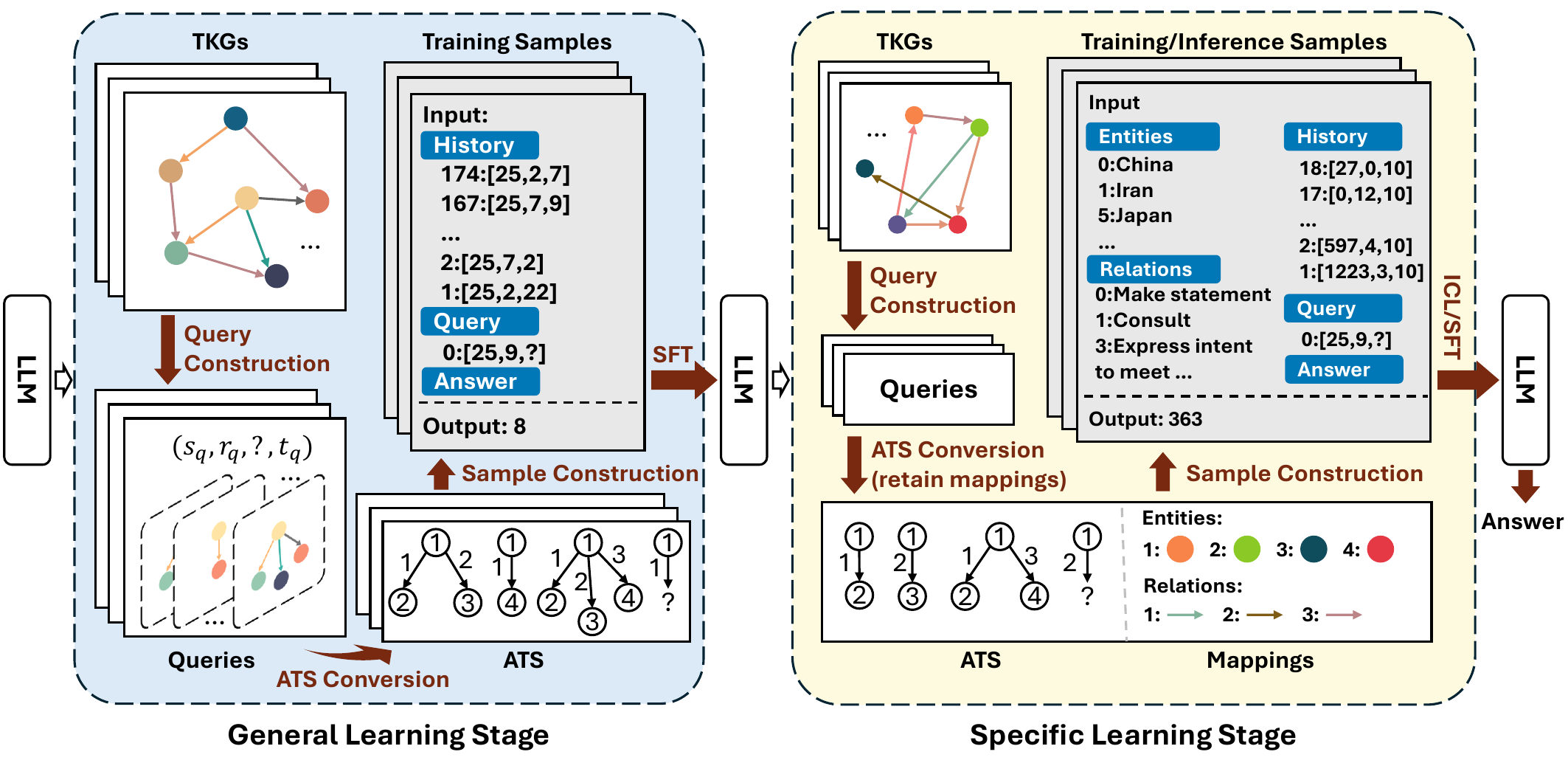}
  \caption{
    The proposed G2S learning framework. 
    Here, ``ATS'',  ``SFT'' and ``ICL'' denote the anonymous temporal structures, 
    supervised fine-tuning, and in-context learning, respectively,
    and each color in KG denotes a specific entity or relation.
  }
  \label{fig:model}
  \vspace{-5mm}
\end{figure*}

To enhance the generalization ability of LLMs on TKG forecasting,
we propose G2S that disentangles the learning process of the general patterns and specific information into two stages,
namely, the general learning and specific learning stages.
Figure~\ref{fig:model} shows the overall diagram of G2S.
We will describe the design of these two learning stages and provide other training and inference details.

\subsection{General Learning Stage}
\label{sec:methodology_gl}
In the general learning stage, there are three main processing steps,
namely, query construction, anonymous temporal structure conversion, and sample construction.
In what follows, we will describe these steps in detail.

\subsubsection{Query Construction} 
In this step, we construct training queries from the TKGs and find historical facts related to these queries.
For each fact $f_q=(s_q,r_q,o_q,t_q)$ in a TKG $\mathcal{G}$, 
we construct two queries $q_1=(s_q, r_q, ?, t_q)$ and $q_2=(?, r_q, o_q, t_q)$ to predict the object and subject entities, respectively.
Then, we select one-hop historical facts $\mathcal{H}_q$ for each query $q$, which is a commonly used history selection strategy in existing approaches~\cite{jin-etal-2020-recurrent,lee-etal-2023-temporal}:
\begin{equation}
    \begin{split}
        \mathcal{H}_{q_1} & = \bigcup_{t<t_q} \{(s_q, r', o', t) \in \mathcal{F}_t\} \\
        \mathcal{H}_{q_2} & = \bigcup_{t<t_q} \{(s', r', o_q, t) \in \mathcal{F}_t\},
    \end{split}
\end{equation}
where $s', o'\in \mathcal{E}$ are arbitrary entities, $r'\in \mathcal{R}$ is arbitrary relation. 
The historical facts are sorted according to their timestamps in ascending order.
Following the convention, we retain the most recent $L$ historical facts.
To better learn general patterns, we collect queries across multiple TKGs with different entities, relations, and time granularities.
For all the TKGs, we adopt the same processing method.

\subsubsection{Anonymous Temporal Structure Conversion}
To eliminate the impact of scenario information in the general learning stage,
we apply an anonymization strategy that converts entities, relations, and timestamps in queries and facts into abstract IDs.

With respect to the timestamps,
previous studies~\cite{lee-etal-2023-temporal,liao-etal-2024-gentkg} utilize the number of timestamps from the start time of each TKG as the ID.
For example, in dataset ICEWS14, the start time is 2014-01-01, and each timestamp is 1 day, so the ID of the timestamp 2014-01-10 is 9.
However, such IDs may create bias between training data and testing data, since testing timestamps are not in the same period as training timestamps.
Therefore, we set the timestamp ID of a query as 0, and the timestamp ID of each historical fact to the number of timestamps from the query time:
\begin{equation}
    \mathcal{A}(t) = t_q - t.
\end{equation}

With respect to entities and relations,
we investigate three different strategies:

\textbf{Frequency ID (FID)} uses the rank of entity frequency (descending order) for each query and corresponding historical facts as ID.
This strategy was proposed in previous studies~\cite{lee-etal-2023-temporal,liao-etal-2024-gentkg}.

\textbf{Global ID (GID)} uses the original ID provided in each dataset.
In this strategy, the ID of each entity and each relation is unique for each TKG.

\textbf{Random ID (RID)} uses the randomly assigned IDs for entities and relations within a query and corresponding historical facts.

Detailed analyses of the advantages and disadvantages of each strategy are described in Section~\ref{sec:anonymization_strategies_analyses}

\subsubsection{Sample Construction}
Based on the anonymous temporal structure, 
we construct training samples to supervised fine-tune the LLM.
Each training sample consists of an input-output pair,
where an input consists of two parts, namely, history and query.
In the history part, each line contains a historical fact in the form of ``$\mathcal{A}(t):[\mathcal{A}(s),\mathcal{A}(r),\mathcal{A}(o)]$'',
where $\mathcal{I}$ denotes the anonymization strategy described above.
The historical facts are listed by their temporal order.
In the query part, the query $(s_q, r_q, ?, t_q)$ is presented in the form of ``$0:[\mathcal{A}(s_q), \mathcal{A}(r_q), ?]$''.
In addition, the input contains an empty answer part at the end to prompt the model to generate the answer.
The output of the training sample is the correct answer entity ID ``$\mathcal{A}(o_q)$''.
For FID and RID strategies, if the answer entity does not occur in historical facts,
it will not be assigned with an ID,
so we set the output to be ``None''.
After being fine-tuned on such training samples, the model captures the general patterns across multiple TKGs.

\subsection{Specific Learning Stage}
\label{sec:methodology_sl}
In the specific learning stage, the processing steps are similar to those presented in Section~\ref{sec:methodology_gl}.
However, there are two key differences:

\textbf{Scenario information mapping.}
In the specific learning stage, the model is required to combine the scenario information with the general patterns to derive more accurate answers.
Therefore, when applying an anonymization strategy to convert entities/relations into IDs, we retain the mappings between entities/relations and IDs.
Then, we add two parts at the beginning of the input, namely, entity and relation parts, according to the mappings.
In the entity part, each line contains the mapping between an entity and its abstract ID in the form ``$\mathcal{A}(e):e$'', as shown in Figure~\ref{fig:model}.
The relation part is done similarly, which contains the mappings between relations and their IDs.

\textbf{Two learning modes.}
The specific learning stage can run in two different modes:

\textit{In-context learning (ICL) mode}.
        In this mode, the model is prohibited from utilizing the training samples.
        Thus, the model can only learn scenario information via in-context learning.

\textit{Supervised Fine-tuning (SFT) mode}.
        In this mode, the model is allowed to utilize the training samples.
        Thus, the model can learn the entities and relations via updating the parameters.

\subsection{Training and Inference Details}
The training objective is to minimize the cross-entropy loss between generated token sequence $O$ and ground-truth token sequence $\hat{O}$, i.e., $\mathcal{L} = CE(O, \hat{O})$.

We use LLaMA3-8B as the backbone model and adopt Low-Rank Adaptation (LoRA)~\cite{hu2022lora} method to effectively fine-tune it.

In the inference phase, considering the efficiency, we only require the model to generate one step.
Then, we adopt the generation score of the tokens provided by the LLM as the ranking score:
\begin{equation}
    \text{score}(o) = \Pr(o | Input),
\end{equation}
where $o$ is a generated token.
Finally, we retain the top 10 valid predictions after filtering the duplicate predictions.

\section{Experimental Setup}

\subsection{Datasets}
We use 5 widely-used datasets in this paper:
ICEWS14~\cite{li-etal-2021-temporal}, 
ICEWS18~\cite{jin-etal-2020-recurrent}, 
YAGO~\cite{mahdisoltani-etal-2013-yago3}, 
GDELT~\cite{jin-etal-2020-recurrent}, 
and WIKI~\cite{leblay-chekol-2018-deriving}.
The basic statistics of the datasets are shown in Appendix~\ref{sec:dataset_statistics}.
Especially, ICEWS14, ICEWS18, and GDELT follow CAMEO~\footnote{https://eventdata.parusanalytics.com/data.dir/cameo.html} schema,
which means relation types in the three datasets are almost the same.
YAGO and WIKI are constructed from open knowledge graphs of the same name, and thus, they have their own schema definitions.

Among these datasets, the training parts of GDELT and WIKI are used in the general learning stage, 
while ICEWS14, ICEWS18, and YAGO are used in the specific learning stage.
The proposed model is evaluated on the later three datasets.
We sample 100,000 training samples from GDELT and 30,000 training samples from WIKI,
approximately according to the ratio of their total training facts.

\subsection{Metrics}
We employ widely used H@1/3/10 to evaluate the proposed model.
These metrics represent the proportion of queries that the model ranks the correct answer entity at top 1/3/10.
Moreover, we adopt the time-aware filtering setting, 
where the valid entities except the correct answer entity are removed from the predictions.
Mean Reciprocal Rank (MRR) is another metric that has often been used in previous studies.
However, this metric is not applicable for LLMs, since LLMs can not produce the full rank of all entities for a query.

\subsection{Baselines}
We compare the proposed method with several baseline methods.
The baseline methods can be divided into two different categories,
i.e., traditional methods and LLM-based methods.
With respect to traditional methods, 
we choose RE-GCN~\cite{li-etal-2021-temporal}, xERTE~\cite{han2021explainable},
TANGO~\cite{han-etal-2021-learning-neural},TITer~\cite{sun-etal-2021-timetraveler}, and TLogic~\cite{Liu_Ma_Hildebrandt_Joblin_Tresp_2022}.
With respect to LLM-based methods, we choose in-context learning~\cite{lee-etal-2023-temporal} method and GenTKG~\cite{liao-etal-2024-gentkg}.
For in-context learning method, we use GPT-NeoX-20B~\cite{black-etal-2022-gpt}, Llama2-7B~\cite{touvron2023llama}, and Llama3-8B~\cite{llama3modelcard} as backbone models, 
noted as GPT-NeoX-ICL, Llama2-ICL, and Llama3-ICL, respectively.

\subsection{Experiment Settings}
We evaluate our model under three different settings:
(1) \textbf{Standard setting}. Under this setting, all training samples can be used to fine-tune the model in the specific learning stage.
(2) \textbf{Zero-shot setting}. Under this setting, no training samples can be used to fine-tune the model in the specific learning stage.
(3) \textbf{Low-resource setting}. Under this setting, the earliest 5\%/20\%/50\% training samples can be used to fine-tune the model in the specific learning stage.

\begin{table*}
    \centering
    \begin{tabular}{lccccccccc}
        \toprule
        \multirow{2}{*}{\diagbox{Model}{Dataset}} &
        \multicolumn{3}{c}{ICEWS14} &
        \multicolumn{3}{c}{ICEWS18} &
        \multicolumn{3}{c}{YAGO} \\
        \cmidrule(lr){2-4} \cmidrule(lr){5-7} \cmidrule(lr){8-10}
         & H@1 & H@3 & H@10 & H@1 & H@3 & H@10 & H@1 & H@3 & H@10 \\
        \midrule
        RE-GCN              & 31.3 & 47.3 & 62.6 & 22.3 & 36.7 & \bf{52.5} & 78.7 & 84.2 & 88.4 \\
        xERTE               & 33.0 & 45.4 & 57.0 & 20.9 & 33.5 & 46.2 & 84.2 & 90.2 & 91.2 \\
        TANGO               & 27.2 & 40.8 & 55.0 & 19.1 & 31.8 & 46.2 & 59.0 & 64.6 & 67.7 \\
        TITer               & 31.9 & 45.4 & 57.5 & 21.2 & 32.5 & 43.9 & 84.5 & 90.8 & 91.2 \\
        TLogic              & 33.2 & 47.6 & 60.2 & 20.4 & 33.6 & 48.0 & 74.0 & 78.9 & 79.1 \\
        \midrule
        llama2-ICL          & 25.8 & 43.0 & 51.0 & 13.5 & 27.6 & 32.6 & 67.7 & 79.0 & 81.8 \\
        llama3-ICL          & 31.9 & 42.4 & 44.1 & 18.1 & 27.2 & 28.8 & 56.9 & 57.9 & 57.9 \\
        GPT-NeoX-ICL        & 32.4 & 46.0 & 56.5 & 19.2 & 31.3 & 41.4 & 78.7 & 89.2 & \bf{92.6} \\ 
        \midrule
        GenTKG              & 36.85 & 47.95 & 53.50 & \bf{24.25} & \bf{37.25} & 42.10 & 79.15* & 83.00* & 84.25* \\
        G2S                 & \bf{38.33}& \bf{54.12} & \bf{68.58} & 23.04 & 35.32 & 46.62 & \bf{87.88} & \bf{90.89} & 90.93 \\
        \bottomrule
    \end{tabular}
    \caption{
        Results of TKG reasoning on datasets ICEWS14, ICEWS18, and YAGO.
        The best results among all models are in boldface.
        The results of GenTKG on YAGO are marked with ``*'' since they use self-developed subset of the standard YAGO dataset. 
        Thus, they are not directly comparable with other methods.
    }
    \label{tab:standard_results}
    \vspace{-3mm}
\end{table*}

\subsection{Implementation Details}
For experiments on all datasets,
the number of historical facts $L$ is set to 50;
context length of prompts is set to 1024;
Learning rate is set to 1e-4;
batch size is set to 8;
number of epochs is set to 1.
We tune the model on the validation set and report the best results.
Unless otherwise specified, G2S utilizes both GDELT and WIKI datasets and adopts RID strategy in general learning stage.
In specific learning stage, G2S adopts GID strategy on ICEWS14 and ICEWS18, and adopts RID strategy on YAGO~\footnote{
    GID strategy is not suitable for YAGO, see Appendix~\ref{sec:multi_token_id_generation_issue} for detailed analyses.
}.
We develop our experiment code based on LlamaFactory~\cite{zheng2024llamafactory} and run the experiments on 4 NVIDIA A800 GPUs.
The general learning stage takes about 2 hours. 
Under the standard setting (fine-tuning on all training samples), specific learning on ICEWS14 takes about 2 hours; 
ICEWS18 takes about 20 hours; YAGO takes about 7 hours.

\section{Results and Analyses}

\subsection{Standard Setting}

To evaluate the performance of G2S when sufficient training data are available,
we conduct experiments under standard settings,
where all training data can be used to fine-tune the model in the specific learning stage. 
The experimental results are shown in Table~\ref{tab:standard_results}.
The methods are split into three groups, the first group contains the traditional methods,
the second contains the LLM-based methods with in-context learning,
and the third contains the LLM-based methods with supervised fine-tuning.
From the table, we can observe that:

(1) G2S outperforms the baselines on ICEWS14 and YAGO under the H@1 metric,
    and achieves the second best performance on ICEWS18,
    which shows the forecasting ability of G2S.

(2) G2S and GenTKG outperform ICL methods in general,
    implying that supervised fine-tuning methods may better learn datasets than ICL methods if sufficient training data are available.
    
(3) G2S slightly underperforms GenTKG on ICEWS18,
    possibly due to the multi-token ID generation issue in G2S.
    A detailed analysis of this result can be found in Appendix~\ref{sec:multi_token_id_generation_issue}.

\subsection{Zero-shot Setting}
\label{sec:zero_shot}

\begin{table*}
    \centering
    \begin{tabular}{lccccccccc}
        \toprule
        \multirow{2}{*}{\diagbox{Model}{Dataset}} &
        \multicolumn{3}{c}{ICEWS14} &
        \multicolumn{3}{c}{ICEWS18} &
        \multicolumn{3}{c}{YAGO} \\
        \cmidrule(lr){2-4} \cmidrule(lr){5-7} \cmidrule(lr){8-10}
         & H@1 & H@3 & H@10 & H@1 & H@3 & H@10 & H@1 & H@3 & H@10 \\
        \midrule
        Frequency           & 24.3 & 38.7 & 53.2 & 14.1 & 26.5 & 40.9 & 76.6 & 85.9 & \underline{92.1} \\
        llama2-ICL          & 25.8 & 43.0 & 51.0 & 13.5 & 27.6 & 32.6 & 67.7 & 79.0 & 81.8 \\
        llama3-ICL          & 31.9 & 42.4 & 44.1 & 18.1 & 27.2 & 28.8 & 56.9 & 57.9 & 57.9 \\
        GPT-NeoX-ICL        & \bf{32.4} & \bf{46.0} & \underline{56.5} & 19.2 & 31.3 & 41.4 & 78.4 & 89.1 & \bf{92.7} \\ 
        \midrule
        $\text{G2S}_{GL(F)}$            & 31.52 & 44.80 & 55.62 & 19.47 & 31.17 & 43.05 & \underline{81.50} & 87.54 & 90.79 \\
        $\text{G2S}_{GL(F+Map)}$        & \underline{32.13} & \underline{45.56} & \bf{56.59} & 19.43 & \bf{31.87} & \bf{44.52} & 81.00 & 87.05 & 90.76 \\
        $\text{G2S}_{GL(R)}$            & 31.95 & 44.87 & 55.61 & 19.54 & 31.34 & 43.29 & 80.40 & 87.32 & 90.86 \\
        $\text{G2S}_{GL(R)\;w.\;WIKI}$  & 32.02 & 44.99 & 55.84 & \bf{19.71} & \underline{31.58} & \underline{43.45} & \bf{86.07} & \bf{90.82} & 91.02 \\
        \bottomrule
    \end{tabular}
    \caption{
        Results of zero-shot TKG reasoning on datasets ICEWS14, ICEWS18, and YAGO.
        The best results among all models are in boldface and the second best results are underlined.
    }
    \label{tab:zeroshot_results}
\end{table*}

To evaluate the generalization ability of G2S, we conduct TKG forecasting under the zero-shot setting.
Since the traditional methods can hardly run under the zero-shot settings,
we only compare G2S with LLM-based methods.
In addition, we adopt a simple frequency baseline proposed by \citet{lee-etal-2023-temporal},
which predicts answers according to entity frequency in historical facts.

We compare several variants of G2S to show the impact of different anonymization strategies in general learning stage:
(1) $\text{G2S}_{GL(F)}$ utilizes GDELT and adopts FID strategy;
(2) $\text{G2S}_{GL(F+Map)}$ add mappings between entity/relation and IDs in addition to $\text{G2S}_{GL(F)}$;
(3) $\text{G2S}_{GL(R)}$ utilizes GDELT and adopts RID strategy;
(4) $\text{G2S}_{GL(R)\;w.\;WIKI}$ jointly utilizes GDELT and YAGO, and adopts RID strategy.
In specific learning stage, the anonymization strategies are selected via the performance on validation set.

We exclude the GenTKG from the baselines since, strictly speaking, its generalization setting is not a zero-shot setting.
The differences between the two settings lie in the following aspects:
(1) The training data (ICEWS14) and testing data (ICEWS18) of GenTKG (generalization) originate from the same source but correspond to different years;
and (2) GenTKG adopts logic-rules mined from ICEWS18.

The experimental results are shown in Table~\ref{tab:zeroshot_results}.
From the table, we can observe that:

(1) Llama3-ICL (8B) underperforms GPT-NeoX-ICL (20B) even though llama3 is a more recent model.
These results imply that the scale of the model has a relatively large impact on the generalization ability in this task. 
Therefore, it is somehow unfair to directly compare G2S (8B) to GPT-NeoX-ICL.
However, $\text{G2S}_{GL(R)\;w.\;WIKI}$ still obtains comparable performance with GPT-NeoX-ICL on ICEWS datasets (especially a higher Hit@10 on ICEWS18) and relatively higher performance on YAGO dataset.
Compared to the baselines except GPT-NeoX-ICL, 
$\text{G2S}_{GL(R)\;w.\;WIKI}$ obtains significantly higher performance on all three datasets, 
which demonstrates the strong generalization ability of this model.

(2) $\text{G2S}_{GL(F)}$ underperforms $\text{G2S}_{GL(F+Map)}$ on ICEWS14 and ICEWS18,
while outperforms it on YAGO.
We guess it is because GDELT shares the same schema with ICEWS14 and ICEWS18 (i.e., relation types are almost similar in these datasets),
$\text{G2S}_{GL(F+Map)}$ may learn useful scenario information from GDELT and apply it to ICEWS datasets.
However, this scenario information may not be so helpful to YAGO.
These results verify our motivation that simultaneously learning general patterns and scenario information may affect the generalization ability of the model.

(3) The performance of $\text{G2S}_{GL(F)}$ and $\text{G2S}_{GL(R)}$ are similar,
which inspires us to use the RID strategy in the general learning stage of the overall framework,
since the RID strategy is simpler in practice.

(4) $\text{G2S}_{GL(R)\;w.\;WIKI}$ outperforms $\text{G2S}_{GL(R)}$ on all three datasets, especially on YAGO.
We guess it is because WIKI may contain more useful general patterns for YAGO than GDELT.
However, the improvements on ICEWS14 and ICEWS18 verify our motivation that,
although TKGs may differ in entities, relations, and time granularities,
they still share some general patterns.

\subsection{Low-resource Setting}
\label{sec:low_resource}
\begin{table*}
    \centering
    \begin{tabular}{lccccccccc}
        \toprule
        \multirow{2}{*}{Model} &
        \multicolumn{3}{c}{5\%} &
        \multicolumn{3}{c}{20\%} &
        \multicolumn{3}{c}{50\%} \\
        \cmidrule(lr){2-4} \cmidrule(lr){5-7} \cmidrule(lr){8-10}
         & H@1 & H@3 & H@10 & H@1 & H@3 & H@10 & H@1 & H@3 & H@10 \\
        \midrule
        RE-GCN      & 13.79 & 22.09 & 30.27 & 19.63 & 29.67 & 39.83 & 24.05 & 36.72 & 48.84 \\
        xERTE       &  6.95 & 14.17 & 25.46 & 17.80 & 29.26 & 42.08 & 22.51 & 34.15 & 46.59 \\
        TANGO       & 11.29 & 17.18 & 22.97 & 11.25 & 17.38 & 23.38 & 14.37 & 17.51 & 22.77 \\
        TITer       & 21.06 & 34.78 & 49.10 & 26.69 & 39.42 & 51.78 & 30.05 & 42.82 & 54.74 \\
        TLogic      & 26.03 & 37.42 & 46.50 & 28.72 & 40.48 & 50.71 & 29.84 & 42.40 & 53.37 \\
        GenTKG      & 30.60 & 42.20 & 49.30 & \bf{34.90} & 46.60 & 54.00 & 36.00 & 48.70 & 55.50 \\
        \midrule
        $\text{G2S}_{SL(F)}$  & \bf{33.65} & \bf{45.88} & 55.81 & 34.12 & 46.63 & 56.63 & 35.10 & 48.08 & 58.07 \\
        $\text{G2S}_{SL(R)}$  & 33.22 & 45.75 & 56.08 & 33.70 & 46.29 & 56.24 & 34.12 & 46.50 & 56.44 \\
        $\text{G2S}_{SL(G)}$  & 31.84 & 44.95 & \bf{56.21} & 33.84 & \bf{47.72} & \bf{61.39} & \bf{36.01} & \bf{50.93} & \bf{65.29} \\
        \bottomrule
    \end{tabular}
    \caption{
        Results of low-resource TKG reasoning on ICEWS14.
        The best results among all models are in boldface.
    }
    \label{tab:lowresource_results}
\end{table*}

To evaluate the generalization ability of G2S under the low-resource setting,
where the earliest 5\%/20\%/50\% samples in ICEWS14 are used to train the model,
and all test samples in ICEWS14 are used to evaluate the model.
In this experiment, we use $\text{G2S}_{GL(R)\;w.\;WIKI}$ in Section~\ref{sec:zero_shot} as the initial model for the specific learning stage.
We compare three variants of G2S with different anonymization strategies in specific learning stage:
$\text{G2S}_{SL(F)}$, $\text{G2S}_{SL(R)}$, and $\text{G2S}_{SL(G)}$ represent the variants of G2S that adopt FID, RID, and GID strategy in specific learning stage, respectively.
Other baselines are the same as those in \citet{liao-etal-2024-gentkg}.

The experimental results are shown in Table~\ref{tab:lowresource_results}.
From the table, we can observe that:

(1) All three variants of G2S outperform the baseline methods via training on 5\% training samples.
Compared to GenTKG, which simultaneously learns general patterns and scenario information,
the proposed two-stage learning framework obtains better generalization ability.

(2) When using 5\% training samples, 
$\text{G2S}_{SL(G)}$ underperforms $\text{G2S}_{SL(F)}$ and $\text{G2S}_{SL(R)}$.
As the number of training samples increases,
$\text{G2S}_{SL(G)}$ gradually outperforms the other two.
This observation implies that,
when there are very few training samples, 
the model is unable to learn the static mapping between entities/relations with global IDs.
However, when more training samples are provided,
GID performs more effectively than the other two strategies.

(3) LLM-based methods, such as GenTKG and variants of G2S,
outperform traditional methods with a large margin,
implying that the LLMs generally obtains better generalization abilities than traditional models.
Thus, they require fewer training samples to reach the same performance as traditional models.

\subsection{Ablation Study}
\begin{table}
    \centering
    \begin{tabular}{lll}
        \toprule
        Model                       & H@1 & H@3 \\
        \midrule
        \multicolumn{3}{l}{\textit{Standard Setting}} \\
        G2S                         & 38.33 & 54.12 \\
            \;\; $w/o.\;Ent$        & $\text{35.67}^{2.66\downarrow}$ & $\text{50.24}^{3.88\downarrow}$ \\
            \;\; $w/o.\;Rel$        & $\text{38.00}^{0.33\downarrow}$ & $\text{53.30}^{0.82\downarrow}$ \\
            \;\; $w/o.\;Ent\&Rel$   & $\text{35.69}^{2.64\downarrow}$ & $\text{50.20}^{3.92\downarrow}$ \\
        \midrule
        \multicolumn{3}{l}{\textit{Zero-shot Setting}} \\
        $\text{G2S}_{GL}$           & 32.02 & 44.99 \\
            \;\; $w/o.\;GL$         & $\text{13.67}^{18.35\downarrow}$ & $\text{28.89}^{16.10\downarrow}$ \\
        \midrule
        \multicolumn{3}{l}{\textit{Low-resource Setting}} \\
        $\text{G2S}_{SL}$           & 33.65 & 45.88 \\
            \;\; $w/o.\;GL$         & $\text{32.13}^{1.52\downarrow}$ & $\text{44.97}^{0.91\downarrow}$ \\
        \bottomrule
    \end{tabular}
    \caption{
        Ablation study on ICEWS14 under all three settings.
    }
    \label{tab:ablation_study}
\end{table}

To verify the effectiveness of each design in G2S,
we conduct an ablation study on ICEWS14 under all three settings.
For standard setting, we evaluate three variants of G2S: 
$w/o.\;Ent$ means the mappings between entities and IDs are removed in the specific learning stage;
$w/o.\;Rel$ means the mappings between relations and IDs are removed in the specific learning stage;
and $w/o.\;Ent\&Rel$ means both kinds of mappings are removed in the specific learning stage.
For zero-shot and low-resource settings, 
we evaluate the variants that remove the general learning stage and use llama3 as the initial model in the specific learning stage,
which is denoted as $w/o.\;GL$.
Here, $\text{G2S}_{GL}$ is $\text{G2S}_{GL(R)\;w.\;WIKI}$ in Section~\ref{sec:zero_shot};
and $\text{G2S}_{SL}$ is $\text{G2S}_{SL(F)}$ in Section~\ref{sec:low_resource} that use 5\% training samples.

The experimental results are shown in Table~\ref{tab:ablation_study},
where we can observe that:

(1) G2S outperforms $w/o.\;Ent$, $w/o.\;Rel$, and $w/o.\;Ent\&Rel$,
which shows the effectiveness of scenario information in TKG forecasting.
In addition, the scenario information about entities are usually more important than that about relations.

(2) $\text{G2S}_{GL}$ and $\text{G2S}_{SL}$ outperform corresponding $w/o.\;GL$ models, respectively,
which shows the effectiveness of the general learning stage in enhancing the generalization ability.

\subsection{Analyses on Anonymization Strategies}
\label{sec:anonymization_strategies_analyses}
From the above experimental results, we can conclude that:

As shown in Table~\ref{tab:zeroshot_results}, in the general learning stage, 
the performance of FID ($\text{G2S}_{GL(F)}$) and RID ($\text{G2S}_{GL(R)}$) strategies are comparable under the zero-shot setting.
Both strategies enhance the generalization ability of the model, 
compared to the model that is additionally trained on the scenario information ($\text{G2S}_{GL(F+Map)}$).

In the specific learning stage, the three strategies show different performances under each setting.
As shown in Table~\ref{tab:standard_results},
GID shows the best performance under the standard setting,
where sufficient samples are available to learn the static mapping between an entity/relation and a unique ID.
However, the different results on 5\% and 50\% training data shown in Table~\ref{tab:lowresource_results} imply that,
GID requires more training data than the other two strategies.
As shown in Tables~\ref{tab:zeroshot_results} and \ref{tab:lowresource_results},
FID performs best under the low-resource and zero-shot settings.
We guess the reason is that FID contains the frequency features in its IDs,
so it outperforms RID.

\section{Conclusions}
This paper investigates the generalization ability of LLMs for temporal knowledge graph forecasting.
Firstly, We identified two kinds of entangled knowledge in TKGs, namely, general patterns and scenario information.
Specifically, we adopted several anonymization strategies to convert the entities and relations in TKGs into abstract IDs.
Via these structures, the LLMs are able to learn the general patterns more comprehensively.
Further, we proposed a general-to-specific learning framework (G2S),
which disentangles the learning of these two kinds of knowledge into two learning stages:
the general and specific learning stages.
To verify the generalization ability of the learned G2S model,
we conducted experiments under standard, zero-shot, and low-resource settings.
The results demonstrate that the model is able to outperform the baseline methods under all three settings,
which shows the effectiveness of the proposed method.

\section*{Limitations}
We find the following limitations:

(1) This paper only studies the generalization ability of G2S on 5 TKGs about geopolitics and general domains.
The generalization ability of the model in other specific domains requires further exploration.

(2) Previous studies have proposed different strategies to select relevant historical facts for queries,
including rule-based and reinforcement learning-based strategies.
In this paper, we only investigate the generalization ability under one-hop strategy for simplicity.
Investigating the generalization ability under other history selection strategies and multi-hop facts will be an important direction for future work.

(3) The GID strategy in G2S usually assigns larger IDs for entities, 
which sometimes requires the model to generate a complete ID through multi-step decoding.
Currently, LLM-based TKG forecasting methods, especially those adopting SFT,
cannot deal with this problem well.

\section*{Acknowledgements}
This work is partially funded by the National Natural Science Foundation of China under grant 62306299, 62441229 and 62406308,
the Lenovo-CAS Joint Lab Youth Scientist Project, 
the project under Grants No. JCKY2022130C039, 
and the Strategic Priority Research Program of the CAS under Grants No. XDB0680102.
Thanks to the reviewers for the constructive discussions and suggestions.

\bibliography{custom}

\appendix

\section{Multi-Token ID Generation Issue}
\label{sec:multi_token_id_generation_issue}

G2S is designed to generate single-token IDs,
which may fail on the queries whose answers are multi-token IDs.
This issue, referred to as the multi-token ID generation issue,
particularly occurs when adopting the GID strategy,
since the IDs assigned by this strategy are usually much larger. 
This issue arises in certain cases, potentially affecting the model's overall performance.
As shown in Table~\ref{tab:multi_token_id},
only 15.4\% of the queries in ICEWS14 have an answer with a multi-token ID,
so the impact of this issue is limited.
For ICEWS18, this percentage increases to 40.2\%,
which impacts the overall performance of the model to some extent.
For YAGO, this percentage is 91.8\%,
which means GID will fail on most queries in this dataset.

\begin{table}[h]
    \centering
    \begin{tabular}{lrrr}
        \toprule
                    & ICEWS14   & ICEWS18   & YAGO \\
        \midrule
        \# Queries  & 17,028    & 91,990    & 39,046 \\ 
        \# MT IDs   & 2,621     & 36,981    & 35,852 \\
        Percentage  & 15.4\%    & 40.2\%    & 91.8\% \\
        \bottomrule
    \end{tabular}
    \caption{
        Statistics on percentage of multi-token answers in validation set of ICEWS14, ICEWS18, and YAGO.
        ``\# Queries'' denotes the number of total queries;
        and ``\# MT IDs'' denotes the number of queries where the answers are multi-token IDs.
    }
    \label{tab:multi_token_id}
\end{table}

\section{Dataset Statistics}
\label{sec:dataset_statistics}
The basic statistics of the datasets are shown in Table~\ref{tab:dataset}.

\section{Additional Results}
Table~\ref{tab:additional_results} shows experimental results which compare G2S with additional baseline models under standard setting.
The baseline models includes TiRGN~\cite{ijcai2022p0299}, CENET~\cite{Xu_Ou_Xu_Fu_2023}, RETIA~\cite{10184853}, L2TKG~\cite{zhang-etal-2023-learning-latent}, and RPC~\cite{10.1145/3539618.3591711}.
As shown in the table, G2S obtains higher or competitive performance compared to the baselines.

\begin{table*}
    \centering
    \begin{tabular}{lccccccccc}
        \toprule
        \multirow{2}{*}{\diagbox{Model}{Dataset}} &
        \multicolumn{3}{c}{ICEWS14} &
        \multicolumn{3}{c}{ICEWS18} &
        \multicolumn{3}{c}{ICEWS05-15} \\
        \cmidrule(lr){2-4} \cmidrule(lr){5-7} \cmidrule(lr){8-10}
         & H@1 & H@3 & H@10 & H@1 & H@3 & H@10 & H@1 & H@3 & H@10 \\
        \midrule
        RiRGN               & 33.73 & 49.85 & 64.46 & 23.10 & 37.90 & 54.20 & 39.07 & 55.75 & 70.11 \\
        CENET               & 30.18 & 41.79 & 55.20 & 23.47 & 34.05 & 47.27 & 38.14 & 49.89 & 62.53 \\
        RETIA               & 32.47 & 48.01 & 63.64 & 21.96 & 36.18 & 51.95 & 34.02 & 49.64 & 64.42 \\
        L2TKG               & 35.36 & - & \bf{71.05} & 22.15 & - & 55.04 & 41.86 & - & 80.69 \\
        RPC                 & 34.87 & 49.80 & 65.08 & \bf{24.34} & \bf{38.74} & \bf{55.89} & 39.47 & 57.11 & 71.75 \\
        \midrule
        G2S                 & \bf{38.33} & \bf{54.12} & 68.58 & 23.04 & 35.32 & 46.62 & 41.39 & \bf{58.15} & \bf{73.34} \\
        \bottomrule
    \end{tabular}
    \caption{
        Results of TKG reasoning on datasets ICEWS14, ICEWS18, and ICEWS05-15.
        The best results among all models are in boldface.
        The experimental results of baselines are cited from \citet{chen-etal-2024-natural}.
    }
    \label{tab:additional_results}
\end{table*}

\begin{table*}
    \centering
    \begin{tabular}{ccrrrrrc}
        \toprule
        Dataset & Schema    & \# Entity  & \# Relation & \# Train     & \# Valid     & \# Test      & Granularity \\
        \midrule
        ICEWS14 & CAMEO     & 7,128     & 230       & 74,845    & 8,514     & 7,371     & 1 day \\
        ICEWS18 & CAMEO     & 23,033    & 256       & 373,018   & 45,995    & 49,545    & 1 day \\
        ICEWS05-15 & CAMEO  & 10,488    & 251       & 368,868   & 46,302	& 46,159    & 1 day \\
        YAGO    & YAGO      & 10,623    & 10        & 161,540   & 19,523    & 20,026    & 1 year \\
        GDELT   & CAMEO     & 7,691     & 240       & 1,734,399 & 238,765   & 305,241   & 15 min \\
        WIKI    & Wikidata  & 12,554    & 24        & 539,286   & 67,538    & 63,110    & 1 year \\
        \bottomrule
    \end{tabular}
    \caption{
        Statistics of the datasets. ``\# Entity/Relation'' denotes the number of entities/relations,
        and ``\# Train/Valid/Test'' denotes the number of facts in train/valid/test set, respectively.
    }
    \label{tab:dataset}
\end{table*}

\section{Prompt Example and Case Study}

\begin{figure}[t]
  \centering
  \includegraphics[width=0.7\columnwidth]{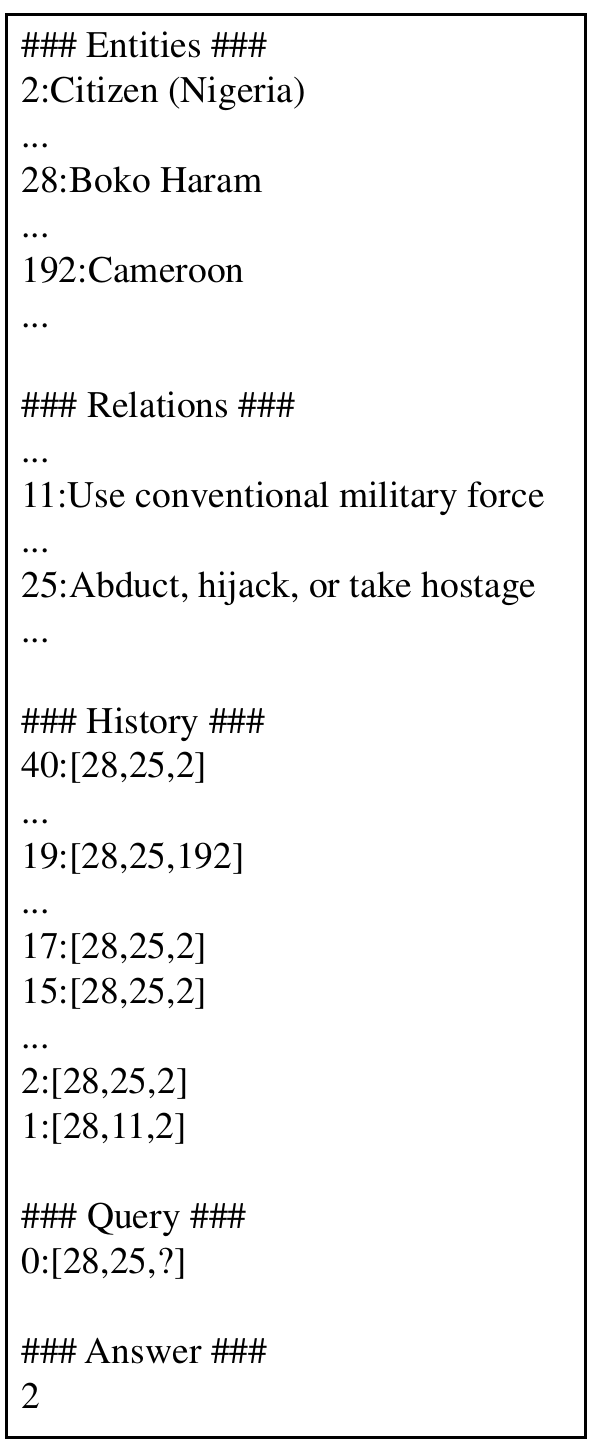}
  \caption{
    An example of the prompt together with answer in specific learning stage.
  }
  \label{fig:prompt}
\end{figure}

An example of the prompt in specific learning stage is shown in Figure~\ref{fig:prompt}.
The prompt in general learning stage excludes ``Entity'' and ``Relation'' parts.

\begin{figure}[t]
  \centering
  \includegraphics[width=0.7\columnwidth]{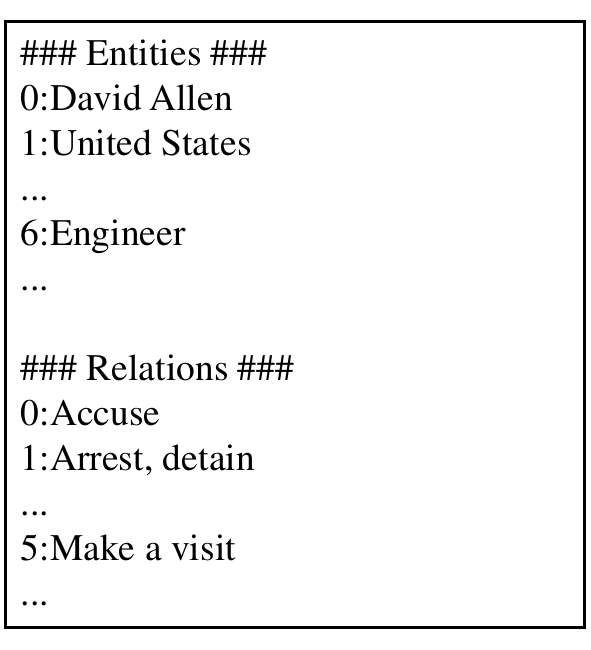}
  \caption{
    The corresponding entity and relation ID mapping.
  }
  \label{fig:example2}
\end{figure}

We found a training query in GDELT (in RID) which is similar in structure but different in semantic,
as shown in Figure~\ref{fig:example1}.
The specific scenario information for this query is shown in Figure~\ref{fig:example2}:

\begin{figure}[t]
  \centering
  \includegraphics[width=0.7\columnwidth]{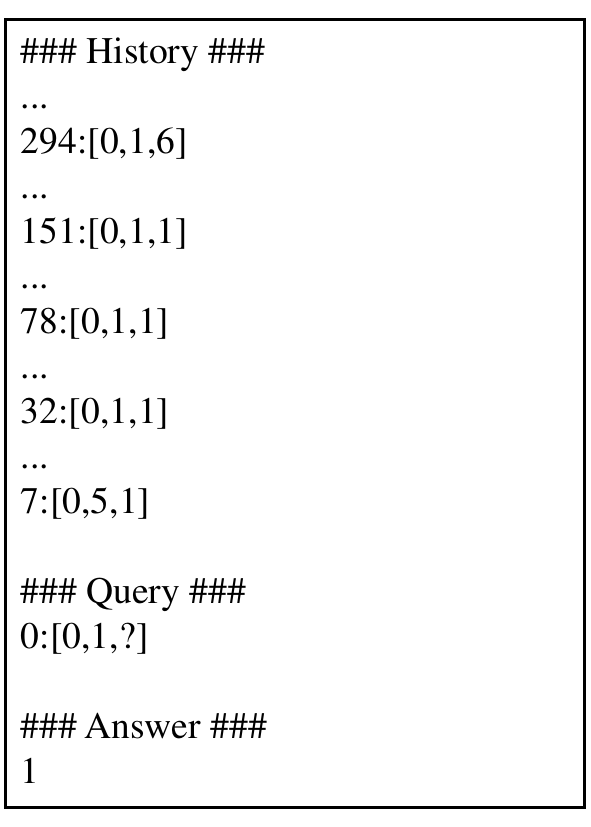}
  \caption{
    A training sample in general learning stage.
  }
  \label{fig:example1}
\end{figure}

Note that the “history” part of the above two queries are represented in anonymous temporal structures, 
thus the model can mine the patterns more easily. 
Though these two samples use different anonymization strategies and are different in time granularity, 
they follow the same pattern. 
This pattern may contain facts related to the query ([28,25, 192], [28,25,2], [0,1,6] and [0,1,1]) and the correct answer depends on the frequency of tail entity. 
This case shows the possibility of learning general patterns across datasets. 

\end{document}